\documentclass[10pt,twocolumn,letterpaper]{article}
\pdfoutput=1 % Eric added per arXiv instructions
\usepackage[pagenumbers]{cvpr} % To force page numbers, e.g. for an arXiv version

\usepackage{graphicx}
\usepackage{amsmath}
\usepackage{amssymb}
\usepackage{booktabs}
\usepackage{multirow}
\usepackage{enumitem}

\usepackage[pagebackref,breaklinks,colorlinks]{hyperref}

\usepackage[capitalize]{cleveref}
\crefname{section}{Sec.}{Secs.}
\Crefname{section}{Section}{Sections}
\Crefname{table}{Table}{Tables}
\crefname{table}{Tab.}{Tabs.}

\def\cvprPaperID{} % *** Enter the CVPR Paper ID here
\def\confName{}
\def\confYear{}

\begin{document}

\title{Training and challenging models for text-guided fashion image retrieval}

\author{Eric McVoy Dodds ~~~~ Jack Culpepper ~~~~ Gaurav Srivastava \\
Yahoo Research \\
{\tt\small \{edodds, jackcul, gaurav.srivastava\}@yahooinc.com}
}
\maketitle

%%%%%%%%% ABSTRACT
\begin{abstract}
Retrieving relevant images from a catalog based on a query image together with a modifying caption is a challenging multimodal task that can particularly benefit domains like apparel shopping, where fine details and subtle variations may be best expressed through natural language. We introduce a new evaluation dataset, Challenging Fashion Queries (CFQ), as well as a modeling approach that achieves state-of-the-art performance on the existing Fashion IQ (FIQ) dataset. CFQ complements existing benchmarks by including relative captions with positive and negative labels of caption accuracy and conditional image similarity, where others provided only positive labels with a combined meaning. We demonstrate the importance of multimodal pretraining for the task and show that domain-specific weak supervision based on attribute labels can augment generic large-scale pretraining. While previous modality fusion mechanisms lose the benefits of multimodal pretraining, we introduce a residual attention fusion mechanism that improves performance. We release CFQ and our code\footnote{\url{https://github.com/yahoo/maaf}} to the research community.
\end{abstract}

%%%%%%%%% BODY TEXT

\section{Introduction} \label{sec:intro}
\begin{figure}
    \centering
    \includegraphics[width=.49\textwidth]{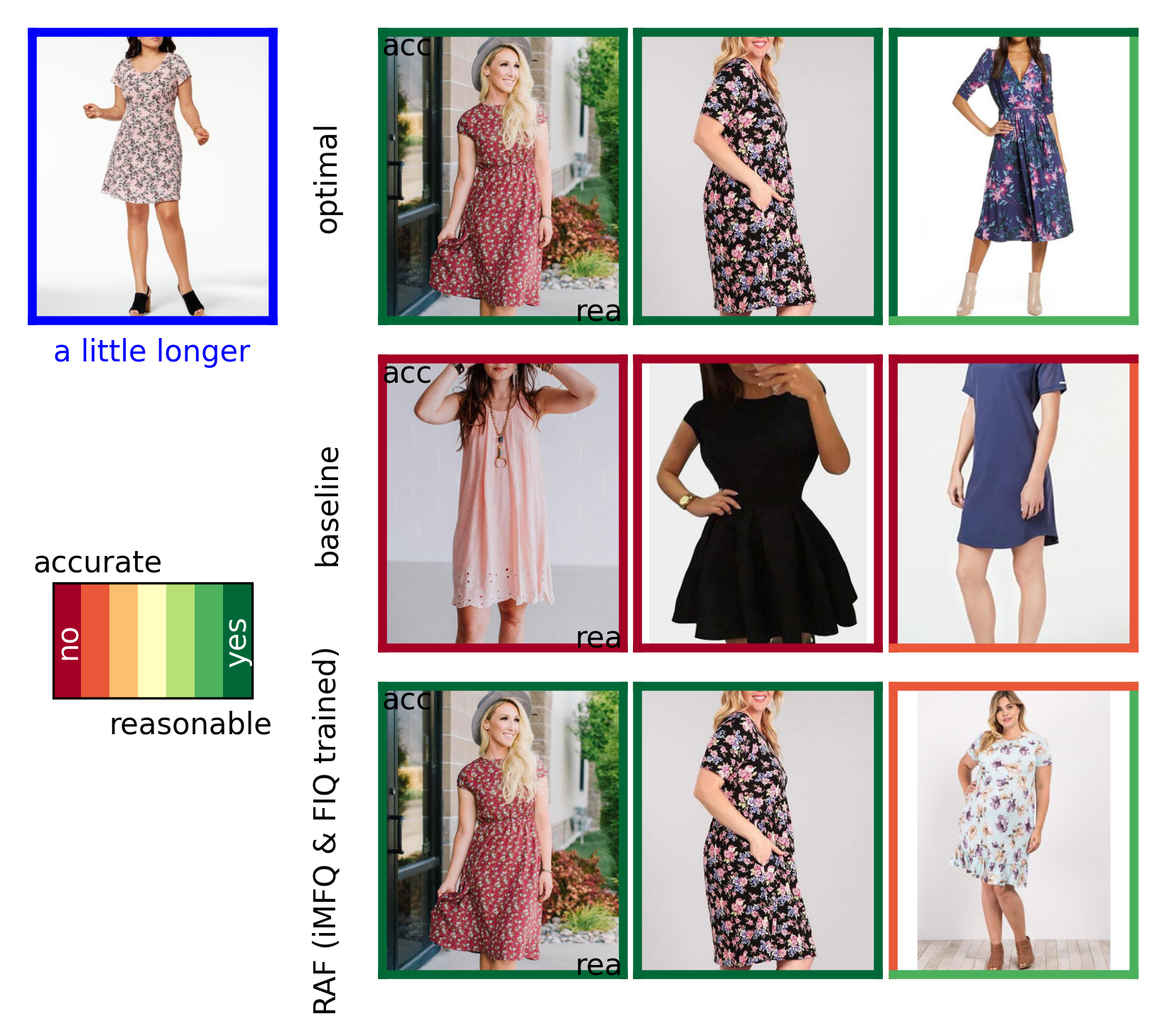}
    \caption{Example query image and caption (blue border and text) and catalog images from our CFQ data: the most relevant images (top) and the top-ranked images retrieved by the baseline CLIP VA model and an RAF model trained on iMFQ and FIQ (see sections \ref{sec:data} and \ref{sec:modeling}). Top-left (bottom-right) corner color indicates average ground-truth accuracy (reasonableness) judgment.}
    \label{fig:cfq_example}
\end{figure}

Fine-grained image retrieval with natural language modification offers the possibility of a flexible, intuitive user experience for search and discovery. Flexible natural language inputs are especially valuable in domains like fashion where items may differ in subtle ways that are not easily captured by categories and lists of attributes alone. The challenge of building a machine learning system to fulfill this promise has recently attracted attention, instigating efforts to collect appropriate data~\cite{vo2019composing, guo2018dialog, Wu_2021_CVPR}. 

Fashion IQ~\cite{Wu_2021_CVPR} contains pairs of product images together with human-written relative captions that can be used to train and evaluate models, but this approach is limited in two ways. First, the lack of negative labels and the noisiness of the positive labels make the evaluation less meaningful. Second, it is impractical to collect enough labels for the model to learn a rich vocabulary of variations from this data alone.

In this paper, we address the first limitation explicitly by collecting an evaluation dataset with positive \textit{and negative} labels over a catalog of possible responses, with judgments made by multiple annotators to decrease noise. Since a fashion assistant system should provide value above and beyond existing systems that search by keywords and filter by attributes, we choose a set of image-text queries that mostly depend on subtle changes from the reference image. We name this dataset Challenging Fashion Queries (CFQ).

We address the scale limitation by studying how to leverage large, readily available datasets.
In particular, we generate relative captions from the attribute labels of the iMaterialist-Fashion dataset~\cite{guo2019imaterialist} to show how such labels can support a text-guided image retrieval model.

We use CLIP~\cite{clip}, a recent model trained on an image-text pair corpus with 400M examples, as a baseline and find that this generalist multimodal pretraining provides an excellent starting point for text-guided fashion image retrieval, even before fine-tuning with images from the apparel domain. Utilizing this model also allows us to take advantage of a language model that has been pretrained on diverse text with a large vocabulary.

While sophisticated mechanisms to fuse text and image features have been shown to improve text-guided image retrieval performance\cite{vo2019composing, Hosseinzadeh_2020_CVPR, Chen_2020_CVPR, maaf}, we show that such mechanisms can actually harm performance when used with CLIP as they disrupt the existing alignment between text and image features. We introduce a fusion approach that does not disrupt this alignment and improves performance beyond the strong CLIP baseline. We also show by explicit ablation that the precise alignment between text and image features drives the strong results we observe.

Our CFQ annotations break down the text-guided image retrieval task in two parts: caption understanding (accuracy) and image similarity conditioned on the captioned change (reasonableness).
These labels allow us to see how different approaches provide different benefits and may trade off between these two parts, and we show that Fashion IQ primarily evaluates models on the accuracy part.

To summarize our contributions:
\begin{enumerate}[wide, labelwidth=!, labelindent=0pt]
    \item Our new Challenging Fashion Queries (CFQ) dataset provides a complementary evaluation to existing text-guided fashion image retrieval data, with both positive and negative labels on the \textit{accuracy} of a relative caption with respect to an image pair and the \textit{reasonableness} of the second image as a suggestion.
    \item We show that a model with multimodal pretraining significantly outperforms the state of the art on the Fashion IQ (FIQ) dataset. Weakly supervised domain-specific pretraining and a modality fusion mechanism tailored to preserve the benefit of multimodal pretraining further improve performance on FIQ.
    \item Several ablation studies and the dual labels of CFQ show the importance of direct alignment of single-modality features and provide insight into the nature of the text-guided image retrieval task and the contribution of each input.
\end{enumerate}
%-------------------------------------------------------------------------
\section{Related Work}

The work of Vo et al. \cite{vo2019composing} established benchmarks and inspired progress on the text-guided image retrieval task by proposing an attention mechanism to fuse image and text information,  superseding several prior methods~\cite{vinyals2015show, noh2016image, santoro2017simple, perez2018film}. Subsequent improvements were made by incorporating dot-product attention into the text-image fusion mechanism~\cite{Hosseinzadeh_2020_CVPR, Chen_2020_CVPR, maaf}, learning image-text compositional embeddings~\cite{chen2020learning}, incorporating a correction module along with fusion~\cite{kim2021dual}, explicitly modeling local versus global image changes~\cite{wen2021comprehensive}, and selecting from and applying corrective comments to a retrieval list, rather than a single retrieved item~\cite{yu2020towards}. Many of these methods are applied to the Fashion IQ dataset and the related Shoes dataset~\cite{guo2018dialog}, which were introduced to study text-guided image retrieval in the context of a dialog recommendation system for apparel. The dataset we introduce in this paper is complementary to Fashion IQ, as we discuss in Section~\ref{sec:data}.

Other relevant fashion datasets include Fashion200k~\cite{han2017automatic} and UT-Zap50k~\cite{yu2014fine, yu2017semantic}, which has been used for a comparison task based on attribute differences. Datasets that have been adapted for text-guided image retrieval include the `States and transformations' dataset~\cite{isola2015discovering}, `Birds to words' \cite{forbes2019neural}, and `Spot-the-diff' \cite{jhamtani2018learning}. Vo et al. synthesized CSS3D, a dataset of simple shapes of various colors, shapes, and sizes~\cite{vo2019composing}.

While our work focuses on a single interaction, others have also studied the setting of a multi-turn back-and-forth with a user~\cite{guo2018dialog, Wu_2021_CVPR, yuan2021conversational}.

\subsection{Vision and language pretraining}
Pretraining of visual-linguistic representations has attracted attention in the context of tasks such as visual commonsense reasoning, visual question answering (VQA), referring expressions, and image retrieval from captions.
Several works~\cite{su2019vl,tan2019lxmert,li2019visualbert,lu2019vilbert,zhou2019unified,alberti2019fusion,chen2019uniter, oscar, sun2019videobert, li2020unicoder, hao2020towards, liu2019aligning} train multimodal Transformer-based~\cite{vaswani2017attention} models in a similar manner to the masked language modeling popularized by BERT~\cite{devlin2018bert} using visuo-linguistic datasets~\cite{sharma-etal-2018-conceptual, chen2015microsoft, ordonez2011im2text, young2014image, hudson2019gqa, goyal2017making, qi2020imagebert}.
This family of Transformer-based models relies on getting visual region proposals from an object detector~\cite{anderson2018bottom, vinvl} and is trained on object annotated datasets~\cite{krishnavisualgenome, lin2014microsoft, kuznetsova2020open, shao2019objects365}.

In contrast, the fashion domain requires understanding fine-grained information in a single clothing items, which some models address by using image regions~\cite{kim2021vilt, gao2020fashionbert, Zhuge_2021_CVPR} instead of object proposals.
Another line of work takes image pixels as input directly, instead of high level region proposals~\cite{huang-seeing-2021, huang2020pixel}. 
Zhuge et al. pretrain vision-language models for fine-grained features in the fashion domain~\cite{Zhuge_2021_CVPR}. However, none of these methods have been applied to text-guided image retrieval.

Liu et al. study text-guided image retrieval in a more general domain and address the incomplete-label problem. They show strong results building on \cite{oscar} but weaker results on the specialized, fine-grained fashion task \cite{liu2021image}.

\subsection{Vision-language contrastive pre-training  and combining different modalities}
% Contrastive Language-Image Pre-training (CLIP)
Radford et al. collected billions of image/alt-text pairs, trained models to match the pairs within large minibatches, and showed that the resulting image models have strong and robust performance on many vision tasks~\cite{clip}. Jia et al. used similar methods with less-filtered data and showed strong results fine-tuning on image classification and vision-language tasks. Of particular relevance to our work, they also showed that their models could be used for text-guided image retrieval~\cite{align}. We have used Contrastive Language-Image Pre-training (CLIP)~\cite{clip} models as our baseline, with vector addition to combine modalities as suggested in~\cite{align}. Other mechanisms for image-text fusion include MAAF~\cite{maaf}, and the utilization of a contrastive loss to align image and text representations before fusing them~\cite{li2021align}. In our work, we experiment with variations on a MAAF-style fusion mechanism.

\subsection{Fine-grained image retrieval}
Several datasets and methods have addressed the challenge of fine-grained image retrieval without language. Examples in the fashion domain include Street-to-Shop~\cite{hadi2015buy}, DeepFashion~\cite{liu2016deepfashion}, and DeepFashion-v2~\cite{ge2019deepfashion2}. The methods that have been applied include the triplet loss~\cite{hoffer2015deep} with sampling techniques~\cite{schroff2015facenet,hermans2017defense} and several variations on cross-entropy~\cite{deng2019arcface,wang2018cosface, movshovitz2017no,sun2020circle,liu2019adaptiveface,he2020softmax}, particularly in the domain of face recognition.

\subsection{Retrieval by attribute changes}
Text-guided image retrieval can be seen as generalizing retrieval based on relative attributes\cite{parikh2011relative} and discrete attribute modifications\cite{zhao2017memory}, since these can be expressed in natural language.

%-------------------------------------------------------------------------
\section{Data} \label{sec:data}
Test data for text-guided fashion image retrieval should serve two purposes: it should allow robust comparisons of different approaches, and it should permit calibration of a model's real-world performance. Existing datasets partly address the first purpose, but the lack of negative labels makes it unclear how metrics will relate to real-world performance. We address this gap and provide additional metrics for comparing approaches by collecting a new dataset we call Challenging Fashion Queries (CFQ)\footnote{CFQ is available for academic research use. For more info see \url{https://webscope.sandbox.yahoo.com/catalog.php?datatype=a&did=92}
}. Since the collection of both positive and negative labels limits the feasible size of the dataset, we turn to larger existing datasets for training by weak supervision to augment training on Fashion IQ, the only dataset with direct supervision for this task.

\subsection{Challenging Fashion Queries (CFQ)}

The CFQ dataset satisfies several desiderata:
\begin{enumerate}[wide, labelwidth=!, labelindent=0pt]
\item CFQ has explicit positive and negative labels for all images in the same category as responses for each query. This allows us to compute precision/recall and other metrics with direct practical significance. Other datasets have only one label per query and assume any other response is negative, which leads to many false negatives as shown in \cite{liu2021image}.\\
\item CFQ has independent labels for accuracy of the caption and for reasonableness of the image pair. These allow us to disentangle these aspects of the task and improve our overall approach as discussed in Section \ref{sec:experiments}.\\
\item CFQ focuses on challenging captions such as relative shape attributes (e.g., ``tighter at the waist"), negations (``not floral"), and subtle holistic changes (``a little less fancy").\\
\item CFQ has minimal noise as each judgment is made by three human annotators and strong disagreements are checked for misunderstanding by two more annotators.
\end{enumerate}

\begin{figure*}
    \centering
    \includegraphics[width=0.8\textwidth]{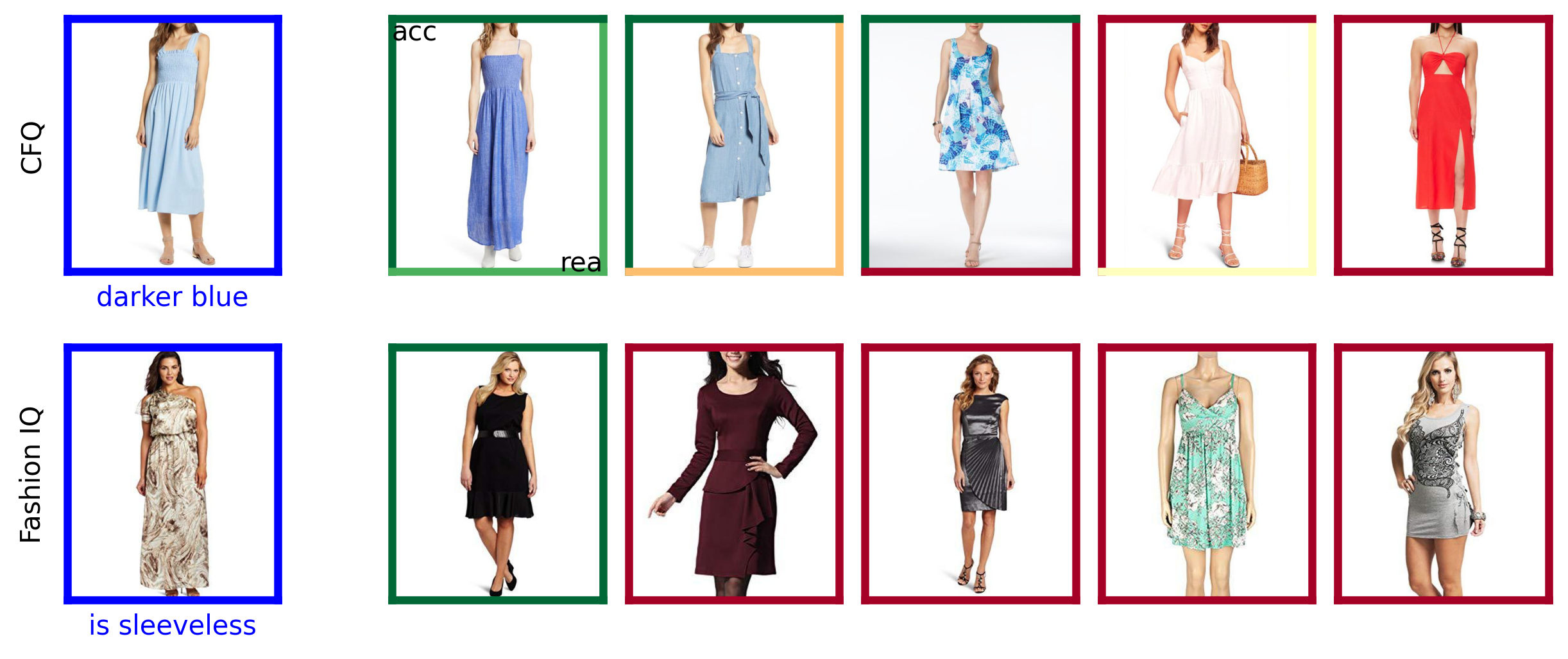}
    \caption{Top: an example CFQ query image and caption, the most relevant catalog image, and four other images selected to show label variety. Bottom: an example FIQ query image and caption, the labeled correct target image, and four random images. Several of these images satisfy the caption but only one is labeled as such.}
    \label{fig:cfq_fiq_comparison}
\end{figure*}

The difference between CFQ labels and Fashion IQ labels is illustrated Fig. \ref{fig:cfq_fiq_comparison}. Note in particular that the lack of complete labels for Fashion IQ means that many assumed negatives are effectively false negatives.
The tradeoff for the desiderata above is that CFQ is relatively small (see Table \ref{tab:dataset-size}). Nevertheless, we found that results are robust and usually repeatable to within a percentage point. See Section \ref{sec:experiments} for results with uncertainties.

\begin{table}
\centering
\begin{tabular}{c|c|c|c}
%\hline
Category     &  Images  & Queries & Judgments \\ \hline
Dress     & 134   & 20 &  2660$\times$3 \\
Gown    & 100   & 10 & 990$\times$3 \\
Sundress & 100  & 10 & 990$\times$3 \\ \hline
Total & 334 & 40 & 4640$\times$3 \\
%\hline
\end{tabular}
\caption{\label{tab:dataset-size} The size of each part of the Challenging Fashion Queries dataset (each judgment is made by three annotators).}
\end{table}

\subsection{Image collection}
We collected a set of images from online advertisements\footnote{Images are not current advertisements and are used for information purposes only. No endorsement is intended.} from various fashion brands and retailers that feature one dress, usually worn by a model. Dresses were chosen as a large category with many subtle variations, where a user may express preferences not easily captured by attribute labels. Since judges found only a few pairs of dresses to be similar enough to make the target a reasonable suggestion given the query item plus a caption, we selected images from finer subcategories for the second half of the queries and the catalogs of response images. Pairs of images from within these subcategories were more likely to be similar than pairs from the broader ``dress" category. The total number of images in each (sub)category is listed in Table \ref{tab:dataset-size}. 

While categories at the level of ``dress" were available with the advertisements, the images in the finer subcategories ``gown" and ``sundress" were chosen by first filtering for the subcategory name in the ad title and then considering examples by hand and including appropriate ones until 100 images were obtained.

\begin{figure}
    \centering
    \includegraphics[width=0.49\textwidth]{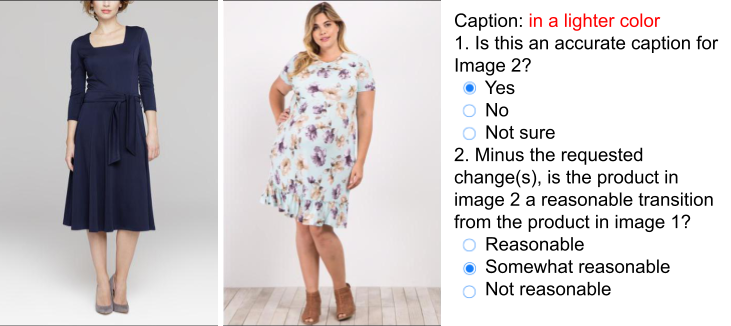}
    \caption{Example of CFQ query image (left), target image(center), and relative caption and judgment (right).}
    \label{fig:cfq_labeling_example}
\end{figure}

For each query-response pair, three annotators were asked two questions as shown in Fig.~\ref{fig:cfq_labeling_example}.
Although for some examples the answers to these questions are clear without further guidance, our focus on challenging queries led to ambiguities of at least two types.

First, natural language generally admits multiple interpretations, particularly with limited context. For example, the caption ``less pink" clearly demands a ``Yes" for question 1 if the response item has some pink coloration but less than the query item, but it is unclear whether a user would be satisfied with an item with no pink at all. To resolve these ambiguities of \textit{linguistic intent}, our annotation team agreed on one consistent interpretation for each caption. For example, annotators were instructed to answer ``No" to question 1 for the caption ``less pink" if the response item had no pink. Our evaluation protocol includes averaging over four phrasings of each relative caption, some more explicit than others.

Second, reasonable people may disagree on what is ``reasonable" for question 2. To mitigate this, the annotation team established criteria for question 2 and reviewed several examples to ensure conceptual agreement.

Counts of each judgment within each of category are shown in Table~\ref{tab:dataset-judgments} along with the fraction of judgments on which all annotators agreed.

\begin{table}
\centering
\begin{tabular}{c|c|c|c|c|c}
%\hline
Question & Judgment & Dress & Gown & Sun & Total  \\ \hline
\multirow{4}{*}{Accurate} & Yes      & 3959  & 1449 & 1461 & 6869  \\ 
         & Not sure & 110   & 23   & 47   & 180   \\ 
         & No       & 3911  & 1498 & 1462 & 6871  \\ 
         & unanim \% &  75.7 & 78.6 & 89.3 & 79.8 \\
\hline
\multirow{4}{*}{Reasonable} & Yes    & 174   & 41   & 89   & 304  \\
         & Somewhat & 1559  & 532  & 451  & 2542 \\
         & No       & 6247  & 2397 & 2430 & 11074 \\
         & unanim \% & 62.5  & 63.6 & 75.4 & 66.0 \\
%\hline
\end{tabular}
\caption{Number of each judgment in each category of CFQ, and the percent of judgments which were unanimous. \label{tab:dataset-judgments}}
\end{table}

\subsection{Evaluation metrics and protocol}\label{sec:metrics}
We use three metrics to summarize a model's performance on CFQ. For each question, we translate the judgments to -1, 0, and +1 and average over annotators to get a collective judgment between -1 and 1 for ``accuracy" (question 1) and another for ``reasonableness" (question 2).

We measure how well a model's scores $s_{q,c}$ for each (query $q$, catalog image $c$) pair predict the answers to the two questions posed to the annotators. In order to compute binary decision metrics we fix a threshold for each question and compute mean (over queries) average precision (mAP). For caption accuracy, we set the threshold at 0. For similarity / reasonableness, we set the threshold as low as possible, \textit{i.e.}, we consider the response image ``reasonable" if at least one annotator judged ``Reasonable" or ``Somewhat reasonable". This low threshold mitigates the problem of class imbalance caused by the annotators' high bar for reasonableness. We also compute an overall mAP metric for overall relevance, where a positive label requires positive answers to both questions.

An example query, the most relevant catalog images, and the top-ranked catalog images according to two models are shown in Fig.~\ref{fig:cfq_example}.

We also evaluate models on what we call FIQ score on the Fashion IQ validation set: the mean of R50 and R10.

\subsection{Weak supervision from attributes}
\begin{figure}
    \centering
    \includegraphics[width=.49\textwidth]{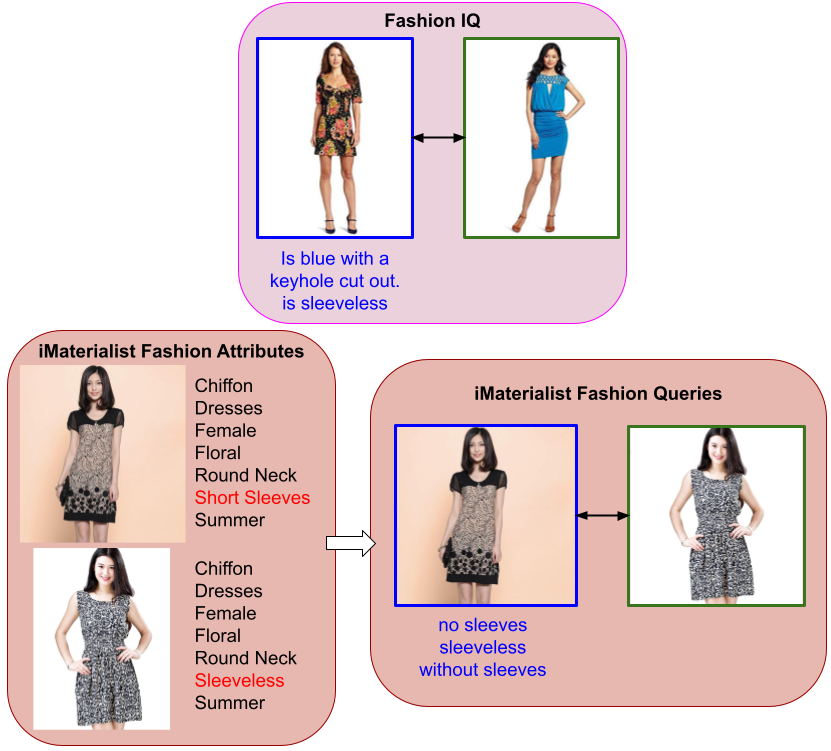}
    \caption{Examples from each of the three training datasets used in this work. Blue border and text indicates a query, green border a labeled positive response.}
    \label{fig:training_data}
\end{figure}

Since it would be very costly to scale the collection of supervised labels for text-guided image retrieval to cover a much larger vocabulary of visual concepts than the 10k-100k image datasets that currently exist, we studied a weakly supervised approach using attribute labels, a type of data that already exists in large quantities. To demonstrate the effectiveness of this approach, we used the iMaterialist Fashion Attribute dataset\cite{guo2019imaterialist}. This dataset contains about 1M images with 228 fine-grained fashion attributes labeled, including groups of attributes for category, gender, color, material, etc. We sampled online from the 152 million pairs of images that differ in their attributes by exactly one label and generated simple relative captions from these differences (e.g., ``black not red"). An example is shown in Fig.~\ref{fig:training_data}. We refer to the dataset together with these pseudo-labels as iMateralist Fashion Queries (iMFQ for short).

While the Fashion IQ dataset \cite{Wu_2021_CVPR} provides tens of thousands of captions with associated image pairs, it has only 2192 distinct words and 448 words that occur at least 5 times in the training set. Without prior knowledge, it is unlikely that a model will learn what a ``wrinkled" garment looks like from the single occurrence of that word in the training set. Our iMFQ labels provide many examples for each attribute as well as important words like ``not". 

Furthermore, paired items in iMFQ share most of their attributes and so are likely to be relatively similar. As we show in Section \ref{sec:experiments}, these pairings provide effective training for the image similarity aspect of the task.

%-------------------------------------------------------------------------
\section{Modeling} \label{sec:modeling}
\begin{figure}
    \centering
    \includegraphics[width=.49\textwidth]{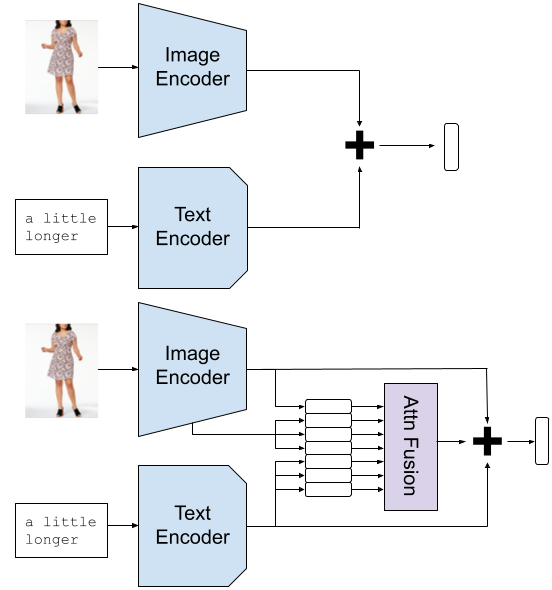}
    \caption{Top: vector addition (VA) of text and image embeddings. Bottom: residual attention fusion (RAF).
    }
    \label{fig:modeling}
\end{figure}

Much prior work on improving the performance of text-guided image retrieval models has focused on the model architecture, especially the mechanism for fusing text and image features~\cite{vo2019composing, Chen_2020_CVPR, maaf, Hosseinzadeh_2020_CVPR, kim2021dual}. For an (image, text) input ($x$, $t$), these models compute a joint image-text embedding
\begin{equation}
    v = f(g_I(x), g_T(t))
\end{equation}
where the image backbone model $g_I$ is pretrained on a visual task such as ImageNet classification and the text backbone model $g_T(t)$ is pretrained separately on a text task such as masked language modeling. The simplest fusion function $f$ is to add embeddings from each modality:
\begin{equation}
    v_\text{VA} = g_I(x) + g_T(t).
\end{equation}
We take as a baseline model the smallest available CLIP model~\cite{clip} together with this vector addition (VA) mechanism. Thus $g_I$ and $g_T$ are respectively an image model and a text model that have been trained to coordinate. 
%is a modified ResNet50 and $g_T$ is a transformer model, and the embedding has dimension 1024. 
As we show in Section \ref{sec:ablations*}, this pretrained cooperation between $g_I$ and $g_T$ leads to strong performance. It is therefore important that any fusion mechanism not disrupt the alignment of the single-modality models. To this end, we introduce \textit{residual attention fusion} (RAF)
\begin{equation}
    v_\text{RAF} = g_I(x) + g_T(t) + \alpha f_\text{AF}([\tilde g_I(x) , \tilde g_T(t)])
\end{equation}
where $f_\text{AF}$ is a Transformer attention\cite{vaswani2017attention} block acting on the concatenation of image and text sequences $\tilde g_I(x)$ and $\tilde g_T(t)$, and $\alpha=0.01$ ensures that the model starts close to the powerful baseline. Several previous works showed state-of-the-art results with a similar mechanism to $f_\text{AF}$ \cite{Chen_2020_CVPR, Hosseinzadeh_2020_CVPR, maaf} alone; here the key is to allow the flexibility of attention fusion while preserving the pretrained alignment of the single-modality embeddings. Thus the model can benefit from CLIP being trained on a huge dataset while also gaining flexibility for the specific task and domain. The VA and RAF approaches are illustrated in Figure~\ref{fig:modeling}. The text sequence for the attention fusion inputs consists of the features corresponding to each token after CLIP's Transformer text model. The image sequence includes the $7\times7$ top-level feature map flattened to 49 vectors, to which we append the ``attention pool" output of CLIP's modified ResNet50 architecture for an image sequence of length 50.

Since we are interested in systems that can retrieve from a sizeable catalog based on a novel query, we evaluate models in a framework where an embedding is computed once for each catalog image and we retrieve from the catalog based on the dot products $s_{q,c}$ between an embedding for a query $q$ and each catalog item $c$. The catalog embeddings are obtained using the same method as the query embeddings but with no text input. All embeddings are normalized, so ranking by dot product is equivalent to using cosine similarity.

Models were trained with batch-wise softmax cross-entropy\cite{sohn2016improved} with a learned inverse-temperature parameter, using Adam~\cite{kingma2014adam} for 14 FIQ epochs and/or 3 iMFQ epochs on one NVIDIA V100 GPU.
%------------------------------------------------------------------------
\section{Experiments} \label{sec:experiments}
We train models on iMFQ and Fashion IQ and report results on Fashion IQ and CFQ. Each experiment is run 5 times and we report the mean and standard deviation of each metric, except for the CLIP model without fine-tuning. 

\subsection{Pretraining and fusion}
Multimodal pretraining is highly effective for the text-guided image retrieval task. Table \ref{tab:comparison_fiq} compares scores for single models on the Fashion IQ validation set as reported in several papers. Multimodal pretraining as provided by CLIP already outperforms the previous state of the art. Our approach enhances this strong result and approaches the score of 52 achieved using ensembles of models for the CVPR 2020 Fashion IQ challenge\footnote{https://sites.google.com/view/cvcreative2020/fashion-iq} with a single model of similar size to the cited methods.

\begin{table}[h]
    \centering
    \begin{tabular}{ll}
    \toprule
Method & FIQ score \\ \midrule
JVSM~\cite{chen2020learning} & 19.26 \\
FiLM*~\cite{perez2018film} & 25.28 \\
%TIRG*~\cite{vo2019composing} & 27.40 \\
Relationship*~\cite{santoro2017simple} & 29.39 \\
CIRPLANT\cite{liu2021image} & 30.20 \\
TIRG**~\cite{vo2019composing} & 31.20 \\
%TRACE w/ BERT~\cite{jandial2020trace} & 34.38 \\ not trustworthy
VAL w/ GloVE~\cite{Chen_2020_CVPR} & 35.38 \\
MAAF~\cite{maaf} & 36.6$^{\pm 0.4}$ \\
CurlingNet***~\cite{yu2020curlingnet} & 38.45 \\
%RTIC~\cite{shin2021rtic} & 38.09 \\
RTIC-GCN~\cite{shin2021rtic} & 39.00 \\
CoSMo~\cite{Lee_2021_CVPR} & 39.45 \\
DCNet~\cite{kim2021dual} & 40.83 \\ 
CLVC-Net~\cite{wen2021comprehensive} & 44.56 \\
\midrule
CLIP\cite{clip} VA & 28.4 \\ 
CLIP VA fine-tuned & 48.1$^{\pm 0.3}$ \\
% CLIP VA fine-tuned text only & 38.4$^{\pm0.1}$ \\
CLIP + iMFQ RAF (Ours) & \textbf{50.2}$^{\pm0.3}$ \\
    \bottomrule
    \end{tabular}
    \caption{Comparison to published work on the Fashion IQ benchmark. * denotes results reported in~\cite{Chen_2020_CVPR}. ** denotes results reported by~\cite{maaf}. *** denotes results reported by~\cite{kim2021dual}. Other results are reported in their respective papers. Methods above the line used only single-modality pretraining. }
    \label{tab:comparison_fiq}
\end{table}

There are two elements to our approach that improve the model's performance on Fashion IQ, as shown in Table \ref{tab:main}. First, weak supervision adapted from attribute labels on a domain-specific dataset (\textit{i.e.,} iMFQ) serves as an effective secondary pretraining before final fine-tuning on Fashion IQ, although training on iMFQ alone actually lowers Fashion IQ performance. Results on our CFQ dataset show why this happens. Training on iMFQ effectively improves a model's ability to tell when an image pair is ``reasonable" but does not similarly improve the model's understanding of relative captions. So optimizing for iMFQ alone is not enough to improve both aspects of the task, and iMFQ-trained models tend to trade some caption understanding for image similarity performance.  This tradeoff for iMFQ-trained models persists after fine-tuning on Fashion IQ.

We find that using a more complex fusion mechanism than vector addition (VA) can improve performance somewhat, but naively applying an existing mechanism tends to hurt performance relative to VA. Table \ref{tab:main} shows two examples of this, TIRG\cite{vo2019composing} and attention fusion (AF). These mechanisms and others have been shown in prior work to improve the performance of models built on single-modality modules that were pretrained (if at all) separately on single-modality data. Since CLIP trains the single-modality modules to directly align, the image and text embeddings ``live in the same space" and can already be added meaningfully before any training with a fusion mechanism. We study this modality alignment by ablation in Section \ref{sec:ablations*}. Here we note that a fusion mechanism (AF) goes from harmful to helpful when we treat it as an initially small correction to VA. Note also that the residual connection to the final embedding is not enough: setting the residual multiplier $\alpha=1$ in the RAF method leads to worse FIQ scores than leaving out the RAF mechanism entirely.

\begin{table*}[h]
    \centering
\begin{tabular}{cc|l|lll}
\toprule
training &           &  FIQ   & \multicolumn{3}{c}{CFQ mAP scores} \\
data  &  fusion   &  score &         accurate &         reasonable &    relevant     \\
\midrule
CLIP &  VA      &  28.4 &  59.2 &  46.7 &  28.4  \\
\midrule
\multirow{5}{*}{FIQ} & TIRG  &  $44.4^{\pm0.2}$ &  $64.7^{\pm0.5}$ &  $53.3^{\pm1.0}$ &  $36.3^{\pm1.4}$ \\
& AF       &  $46.0^{\pm0.3}$ &  $59.0^{\pm1.1}$ &  $\mathbf{55.7}^{\pm0.5}$ &  $34.7^{\pm1.1}$  \\
& RAF $\alpha=1$ &  $46.4^{\pm0.5}$ &  $59.6^{\pm0.7}$ &  $\mathbf{55.7}^{\pm0.6}$ &  $34.9^{\pm0.5}$ \\
& VA       &  $48.1^{\pm0.3}$ &  $66.8^{\pm0.3}$ &  $52.9^{\pm0.3}$ &  $37.4^{\pm0.3}$ \\
& RAF      &  $49.7^{\pm0.3}$ &  $\mathbf{67.3}^{\pm0.2}$ &  $53.0^{\pm0.6}$ &  $\mathbf{38.0}^{\pm0.5}$ \\
\midrule
\multirow{2}{*}{iMFQ} & VA      &  $24.9^{\pm0.1}$ &  $60.5^{\pm0.9}$ &  $53.2^{\pm0.3}$ &  $33.2^{\pm0.9}$ \\
& RAF      &  $24.4^{\pm0.6}$ &  $60.0^{\pm0.7}$ &  $53.9^{\pm0.4}$ &  $33.4^{\pm0.8}$ \\
\midrule
\multirow{2}{*}{\shortstack{iMFQ\\ \& FIQ}} & VA &  $49.9^{\pm0.3}$ &  $62.9^{\pm0.6}$ &  $\mathbf{55.7}^{\pm0.3}$ &  $37.1^{\pm0.6}$ \\
& RAF &  $\mathbf{50.2}^{\pm0.3}$ &  $63.9^{\pm0.5}$ &  $\mathbf{55.6}^{\pm0.1}$ &  $\mathbf{38.0}^{\pm0.4}$ \\
\bottomrule
\end{tabular}
    \caption{Pretraining on iMFQ improves models' FIQ scores; CFQ metrics shows that these models are better at distinguishing conditionally similar (``reasonable") images but worse at understanding captions. In general our best results are achieved using RAF as the fusion mechanism. All models in this table start from the CLIP checkpoint whose results are reported in the top row. 
    % Fashion IQ (FIQ) and our Challenging Fashion Queries (CFQ) metrics for CLIP-based models trained on FIQ, iMFQ, or sequentially on iMFQ then FIQ. 
    See section \ref{sec:modeling} for fusion mechanism descriptions, and \cite{vo2019composing} for TIRG.}
    \label{tab:main}
\end{table*}

\subsection{Caption understanding and image similarity}
\label{sec:subtasks}

The text-guided image retrieval task has usually been framed in terms of a \textit{modifying caption}, where the target image is like the query image except for a change. While the captions in CFQ were chosen to ensure the query image is relevant (``with less pink"), many FIQ captions make no reference to the query image (``is blue") and the image pairs are not always similar. In fact we find that a CLIP-based model that ignores the query image entirely can achieve performance competitive with the state of the art on FIQ. Table \ref{tab:base-ablations} shows this result (``text trained") and also that a model that instead ignores the caption (``img trained") achieves comparably poor performance. As one might expect, the image-only model achieves decent performance on the CFQ reasonableness mAP and the text-only model does well on accuracy mAP, but each performs poorly on the other metric and therefore on the overall mAP. 

\begin{table}[h]
\begin{tabular}{l|l|lll}
\toprule
{}    & FIQ & \multicolumn{3}{c}{CFQ mAP scores} \\
model & score &          acc &          rea &              relevant      \\
\midrule
random          &   $00.6^{\pm0.1}$ &  $53.2^{\pm0.7}$ &  $40.1^{\pm0.7}$ &  $22.5^{\pm0.8}$  \\
CLIP VA      &  28.4 &  59.2 &  46.7 &  28.4  \\
\midrule
CLIP img &   08.9 &  53.4 &  53.1 &  29.6  \\
CLIP text  &  22.7 &  59.8 &  40.5 &  24.1  \\
\midrule
img trained   &   $9.9^{\pm1.1}$ &  $53.9^{\pm0.5}$ &  $53.8^{\pm0.9}$ &  $30.5^{\pm1.1}$  \\
text trained    &  $38.4^{\pm0.1}$ &  $66.1^{\pm0.4}$ &  $44.0^{\pm0.2}$ &  $30.2^{\pm0.4}$  \\
\bottomrule
\end{tabular}
\caption{Models rely more on captions than on query images. Top: a model that outputs random embeddings, and CLIP with VA and no additional training, for reference. Middle: CLIP img (text) only uses the query image (text) and the target image. Bottom: the models from the middle fine-tuned on FIQ.}
\label{tab:base-ablations}
\end{table}

While a good text-guided image retrieval system should do well at both caption understanding and conditional image similarity, these subtasks may be in tension for actual systems. Table \ref{tab:image-only} shows using the CFQ reasonableness mAP that the models we study do better at predicting reasonableness judgments if the text input is hidden, although the gap is mitigated by training on both iMFQ and FIQ. At the level of performance currently achieved, accounting for the relative caption appears unimportant for conditional image similarity compared to the already challenging problem of matching images of similar products.

\begin{table}[]
    \centering
\begin{tabular}{ll|l|l}
\toprule
& training & \multicolumn{2}{c}{CFQ reasonableness mAP} \\
fusion & data & fused & image only \\
\midrule
VA & CLIP    &   46.7 &   53.1 \\
VA & FIQ       &  $52.9^{\pm0.3}$ &  $56.5^{\pm0.2}$ \\
RAF & FIQ      &  $53.0^{\pm0.6}$ &  $57.6^{\pm0.2}$ \\
VA & iMFQ      &  $53.2^{\pm0.3}$ &  $55.7^{\pm0.3}$ \\
RAF & iMFQ     &  $53.9^{\pm0.4}$ &  $56.0^{\pm0.2}$ \\
VA & iMFQ \& FIQ &  $55.7^{\pm0.3}$ &  $56.7^{\pm0.3}$ \\
RAF & iMFQ \& FIQ &  $55.6^{\pm0.1}$ &  $57.2^{\pm0.0}$ \\
\bottomrule
\end{tabular}
    \caption{Models do better at conditional image similarity if they ignore the text. In the ``image only" column, models were scored without access to the modifying captions.}
    \label{tab:image-only}
\end{table}

\subsection{Modality alignment}
\label{sec:ablations*}

To better understand the strong performance of CLIP applied directly to the text-guided image retrieval task, we conducted an ablation on the alignment between the CLIP modules. Specifically, we intentionally disrupted this alignment either by scrambling the channels of the text embedding or by replacing the text module with another CLIP-trained text module that was not trained in cooperation with the image module. The CLIP authors reported that few-shot training on ImageNet could recover the zero-shot performance using the text model to embed the category labels with only 4 images per category\cite{clip}. In the text-guided image retrieval setting, however, we find that it takes training on the entire Fashion IQ training set to surpass the baseline (see \cref{tab:misalignment}). Furthermore, fine-tuning from ordinary pretrained CLIP provides an enormous benefit (50\% relative improvement in FIQ score) over fine-tuning from a model with disrupted modality alignment. We conclude that it is the direct alignment between the embeddings of the image and text modules of CLIP that is essential to CLIP's performance on this task.

\begin{table}[h]
\setlength{\tabcolsep}{4pt}
    \centering
\begin{tabular}{lllll}
\toprule
        &  FIQ   & \multicolumn{3}{c}{CFQ mAP scores} \\
model   &  score &         accurate &         reasonable &             relevant \\
\midrule
random           &   $0.6^{\pm0.1}$ &  $53.2^{\pm0.7}$ &  $40.1^{\pm0.7}$ &  $22.5^{\pm0.8}$ \\
CLIP VA       &  28.4 &  59.2 &  46.7 &  28.4 \\
\midrule
scramble         &   $3.1^{\pm0.7}$ &  $53.1^{\pm1.0}$ &  $45.4^{\pm1.4}$ &  $25.4^{\pm0.8}$ \\
~~~~trained &  $35.6^{\pm0.3}$ &  $61.5^{\pm3.5}$ &  $48.1^{\pm1.3}$ &  $31.9^{\pm3.2}$ \\
mismatch         &   2.6 &  51.3 &  40.9 &  22.2 \\
~~~~trained &  $32.2^{\pm0.4}$ &  $60.5^{\pm1.4}$ &  $49.3^{\pm1.1}$ &  $32.1^{\pm1.7}$ \\
\bottomrule
\end{tabular}
    \caption{The benefit of CLIP-based models depends on modality alignment. Top: random embeddings, and CLIP with VA, for reference. Bottom: CLIP models with disrupted modality alignment. Without further training, disrupted models are little better than chance. The ``trained" models are disrupted as in the line above and then trained on FIQ for 42 epochs (compared to 14 for our non-disrupted models). We only ``mismatch" CLIP with the one text module, so there is no inter-experiment standard deviation to report.}
    \label{tab:misalignment}
\end{table}

\subsection{Average precision variation across queries}

\begin{figure}
    \vspace{-0.75cm}
    \centering
    \includegraphics[width=0.5\textwidth]{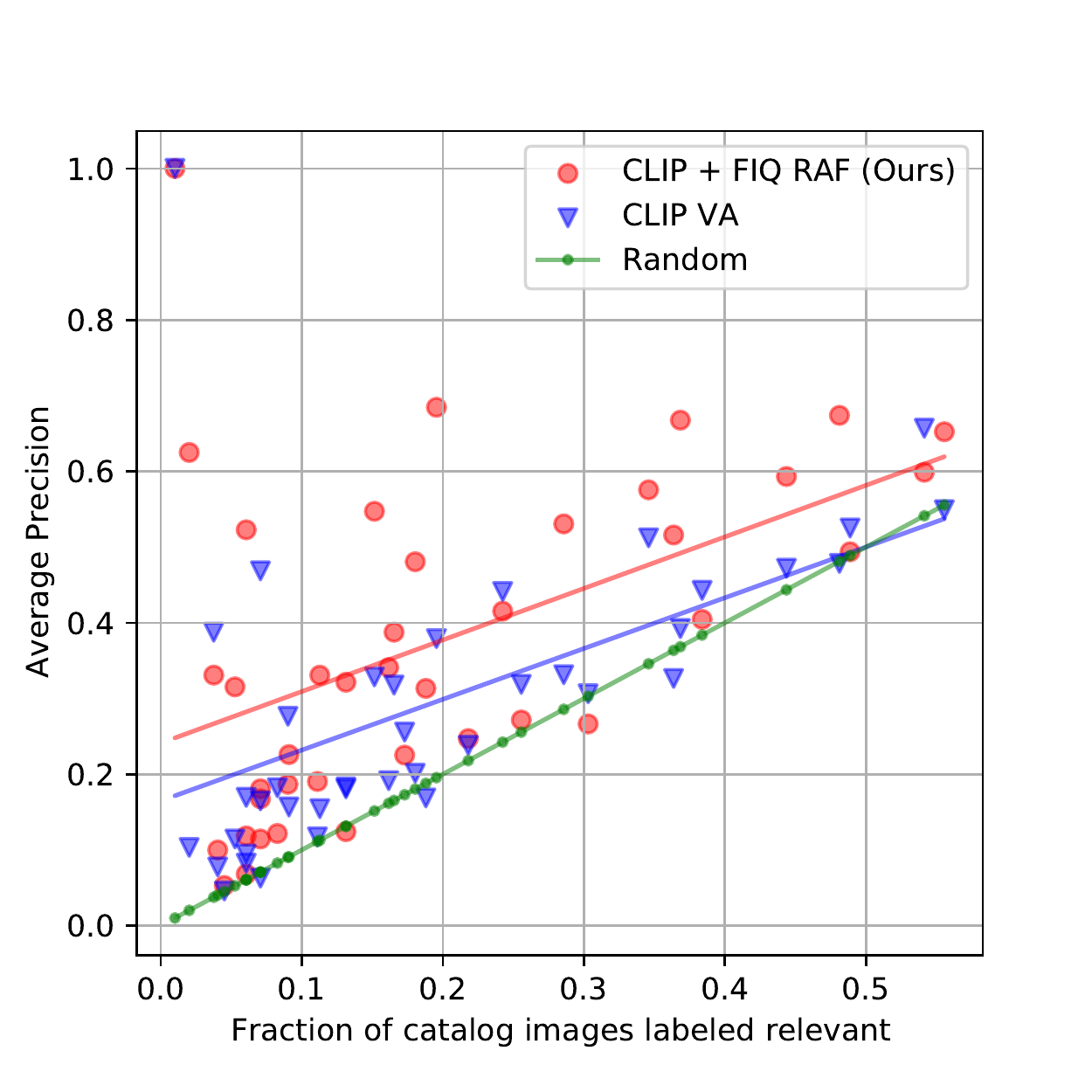}
    \caption{Per query average precision and best fit line for CLIP + FIQ RAF and the baseline model (CLIP VA), compared to the expected AP for a random model. Models show their strength especially on queries with few relevant catalog images.}
    \label{fig:ap-vs-tpr}
\end{figure}

There is significant variability in the number of response images considered relevant overall for each query (see \cref{sec:metrics}). The fraction of relevant responses per query ranges from 0.01 to 0.56, and is the expected AP for a random model. Figure \ref{fig:ap-vs-tpr} shows how two models perform on individual queries, revealing that gains come disproportionately from queries with few relevant catalog images.

\subsection{Types of caption}

\begin{figure}
    \centering
    \includegraphics[width=0.49\textwidth]{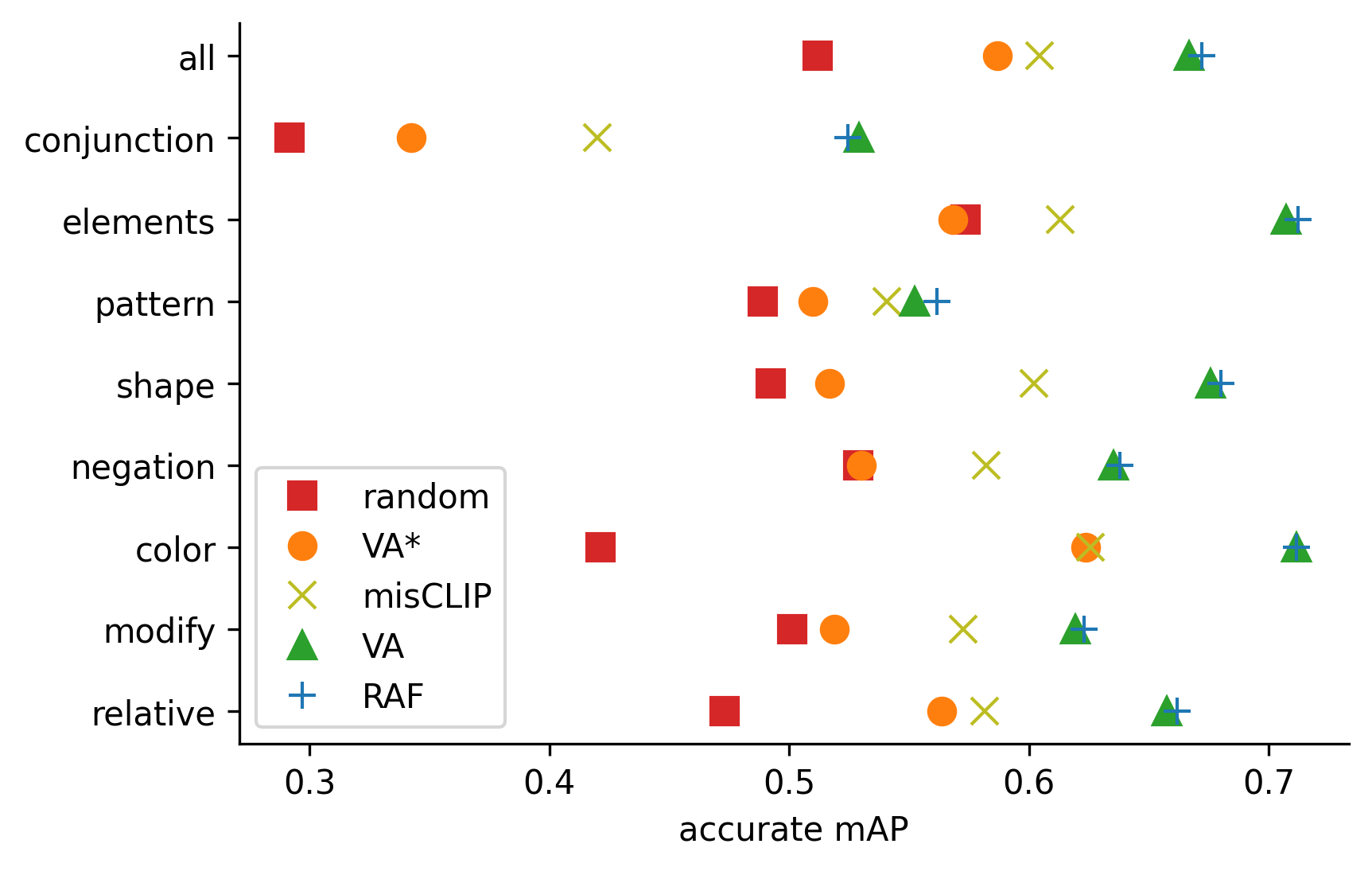}
    \caption{Selected models' CFQ accuracy mAP for several types of caption. All but VA$^\ast$ are fine-tuned on FIQ.}
    \label{fig:attributes}
\end{figure}

Natural language can describe diverse changes, and we grouped the captions in CFQ based on whether they addressed certain attributes (elements, pattern, shape, color), whether they included a conjunction or a negation, and whether they requested a discrete attribute change (c.f. \cite{zhao2017memory} or a change in degree (``relative" changes as in \cite{parikh2011relative}). Fig. \ref{fig:attributes} uses CFQ accuracy mAP within these (not mutually exclusive) groups to show how several model properties improve performance across the caption types. While pretrained modality alignment, FIQ fine-tuning, and use of RAF increase accuracy mAP broadly, we see some variation including a larger improvement for relative changes compared to modifications.

\subsection{Generality}

While we focused on fashion, our modeling approach is appropriate for any domain that emphasizes fine-grained understanding of single objects. To show this generality we report results (with no hyperparameter tuning) on two other datasets that have been used for text-guided image retrieval in Table \ref{tab:extras}.

\begin{table}[h]
    \centering
\begin{tabular}{c|ll|lll}
  &  \multicolumn{2}{c|}{Birds-to-Words}   & \multicolumn{3}{c}{MIT-States} \\
 method &  R@10 &         R@50 &         R@1 &    R@5   & R@10     \\
\hline
MAAF\cite{maaf} & 34.8 & 66.3 &  12.7 & 32.6 & 44.8     \\ %\cite{maaf}
LB\cite{Hosseinzadeh_2020_CVPR} & - & - & 14.7 & 35.3 & 46.6 \\
\cite{zhang2020joint} & - & - & 14.27 & 33.21 & 45.34 \\
RTIC\cite{shin2021rtic} & 37.56 & 67.72 & - & - & - \\
\hline
VA$^\ast$ &    22.8 &  55.9 &  8.7 &  29.0 &  40.8  \\
VA &  33.1 &  57.0 & 12.2 &  35.2 &  48.0 \\
TIRG & 39.8 &  76.6 & \textbf{18.2} &  42.1 &  53.6 \\
AF  & 46.5 &  79.0 & 16.9 &  40.3 &  51.8 \\
RAF & \textbf{51.1} &  \textbf{82.0} & 15.7 &  \textbf{43.7} &  \textbf{56.6} \\
\hline
\end{tabular}
    \caption{Results on Birds-to-Words \cite{forbes2019neural} and MIT States \cite{isola2015discovering}. Top: previous SotA. Bottom: our models as in Table 4. Methods below the divide use CLIP-pretrained modules plus the stated fusion mechanism. $\ast$ no fine-tuning.}
    \label{tab:extras}
\end{table}
%-------------------------------------------------------------------------
\section{Conclusion}
The problem of developing a system that understands and can retrieve high-quality results for a wide variety of subtle visual changes as expressed in natural language poses many challenges. Since a purely supervised approach would be impractical to scale to a large vocabulary of relevant visual concepts, we studied approaches based on using large pretrained models and pretraining with multimodal data readily obtainable from existing fashion catalogs. Our new evaluation dataset Challenging Fashion Queries (CFQ) provides a difficult benchmark showing that our approach is effective but leaves much room for improvement. 

By disrupting the alignment between the single-modality modules of a multimodal-pretrained model, we showed that multimodal pretraining per se -- as opposed to strong pretraining within each modality alone -- is highly beneficial for this task. Indeed, all the other effects we observe and the differences between the various modality fusion mechanisms proposed in the literature (\textit{e.g.}\cite{maaf, kim2021dual}), while significant and interesting, are small by comparison. Future work should build on this observation. Our experiments also show that a properly trained fusion mechanism can improve performance, but it is crucial that the fusion mechanism be compatible with the learned cooperation between modalities from pretraining. Incorporating modality fusion in a large-scale pretraining effort may yield further improvements on text-guided image retrieval and other tasks.

\section{Limitations and Assets}

Although the methods discussed here offer benefits, they cannot fully overcome the limits of real datasets. In particular, even very large and/or carefully constructed pretraining datasets will omit or neglect some visual concepts and will have biases towards or against some styles, groups of people, etc. Addressing these biases directly will be important for a real-world system but is outside the scope of this paper.

The top score we achieve on our CFQ overall relevance mAP is 38.0, suggesting a great deal of room for improvement. While the exhaustive labels of CFQ provide a closer approximation to a test of real-world performance and permit the calculation of useful metrics such as recall at a fixed precision, the absolute values of such metrics would depend on application specifics such as the users, the user interface, and the catalog. CFQ is also limited to a category of items, so any method that singles out that category may achieve deceptively high scores.

\section*{Acknowledgments}
The authors thank the members of the Content Analysis and Knowledge Engineering team at Yahoo for their assistance defining and collecting the Challenging Fashion Queries judgments. We also thank the other members of the Visual Intelligence team at Yahoo for feedback and support throughout our work.

%%%%%%%%% REFERENCES
% \clearpage
{\small
\bibliographystyle{ieee_fullname}
\bibliography{imgtext}
}
\clearpage

\appendix
\section{Limitations and Assets}

Although the methods discussed here offer benefits, they cannot fully overcome the limits of real datasets. In particular, even very large and/or carefully constructed pretraining datasets will omit or neglect some visual concepts and will have biases towards or against some styles, groups of people, etc. Addressing these biases directly will be important for a real-world system but is outside the scope of this paper.

The top score we achieve on our CFQ overall relevance mAP is 38.0, suggesting a great deal of room for improvement. While the exhaustive labels of CFQ provide a closer approximation to a test of real-world performance and permit the calculation of useful metrics such as recall at a fixed precision, the absolute values of such metrics would depend on application specifics such as the users, the user interface, and the catalog. CFQ is also limited to a category of items, so any method that singles out that category may achieve deceptively high scores.

We accessed the following existing datasets:

\begin{enumerate}[wide, labelwidth=!, labelindent=0pt]
\item iMaterialist-Fashion Attributes\footnote{https://github.com/visipedia/imat\_fashion\_comp}\cite{guo2019imaterialist} lists no license but was provided for non-commerical use for a Kaggle competition\footnote{https://www.kaggle.com/c/imaterialist-challenge-fashion-2018/rules}.
\item Fashion IQ\footnote{https://github.com/XiaoxiaoGuo/fashion-iq}\cite{guo2019fashion} was released under the CDLA\footnote{https://cdla.dev/}.
\end{enumerate}

Our code extended the MAAF codebase\footnote{https://github.com/yahoo/maaf, Apache License 2.0}, which used the TIRG codebase\footnote{https://github.com/google/tirg, Apache License 2.0}\cite{vo2019composing} as a starting point. We also adapted code from CLIP\footnote{https://github.com/openai/CLIP, MIT License}\cite{clip}. 

Human judgments for the CFQ dataset were obtained with consent, and no personally identifiable information is included. 

\section{CFQ examples}
\label{apx:data}
\begin{figure*}
    \centering
    \includegraphics[width=0.99\textwidth]{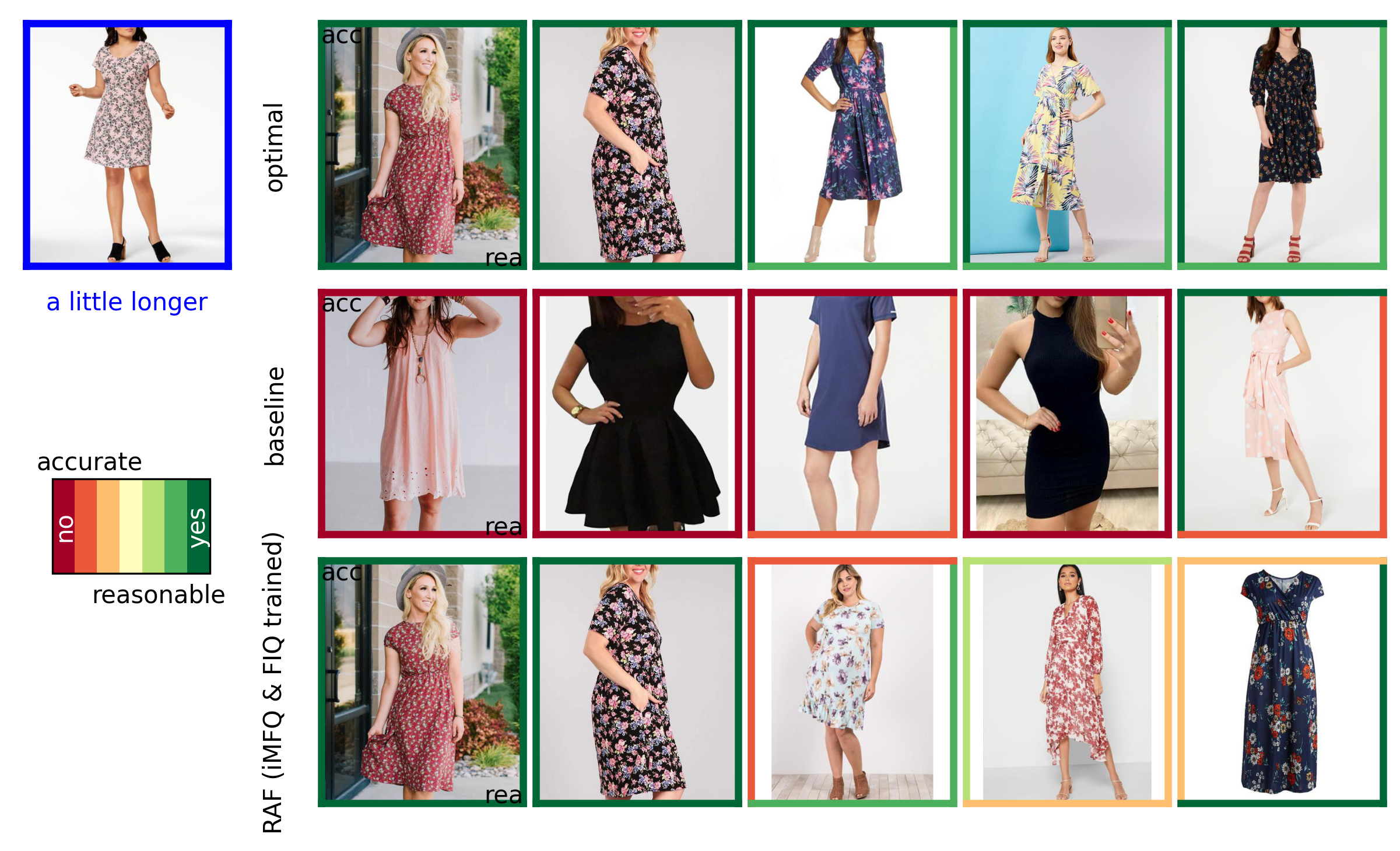}
    \caption{An example from the broad ``dress" category. Baseline AP scores: 44.8 accuracy, 57.3 reasonableness. RAF model AP scores: 64.2 accuracy, 81.3 reasonableness.}
    \label{fig:supp_example_longer}
\end{figure*}

\begin{figure*}
    \centering
    \includegraphics[width=0.99\textwidth]{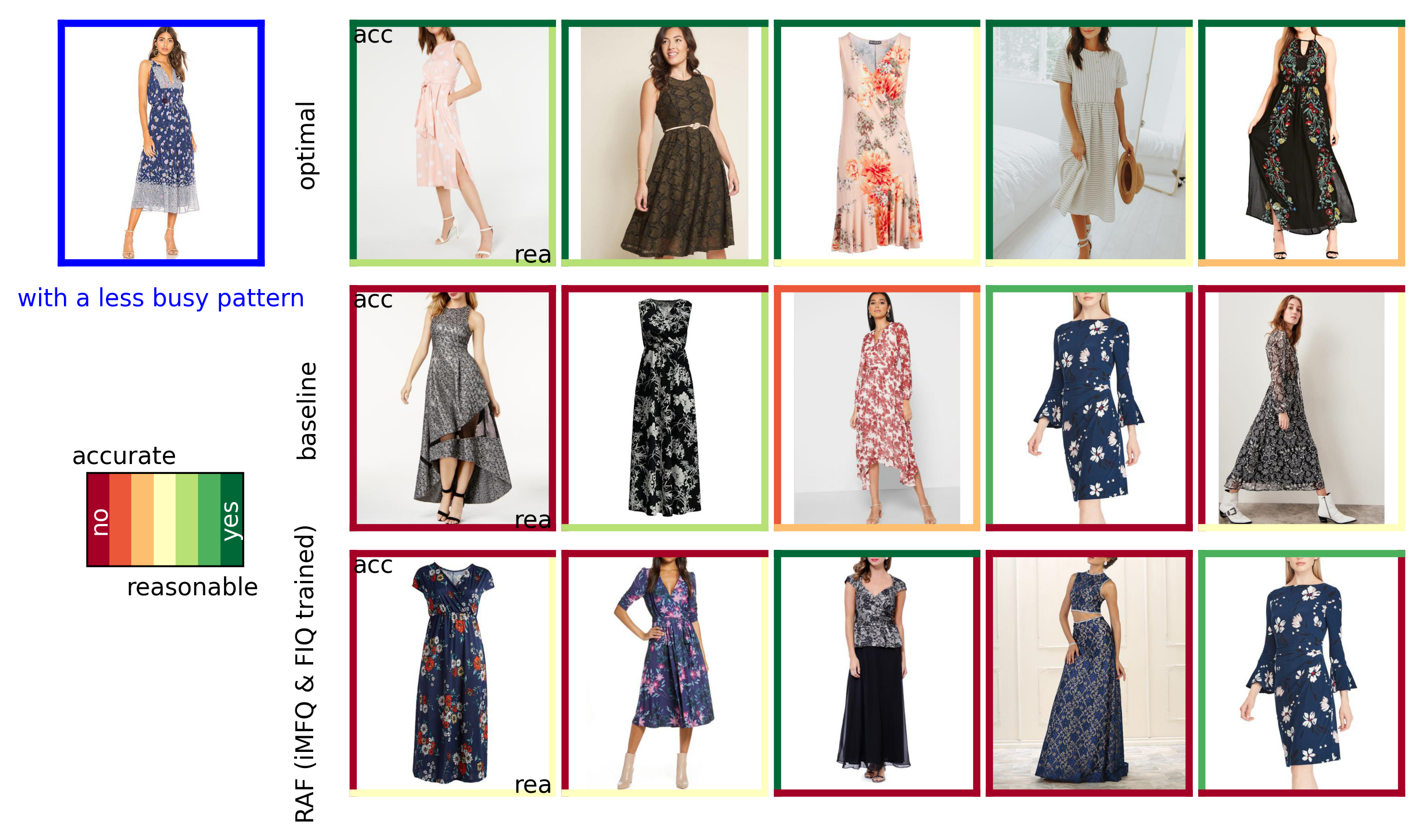}
    \caption{From ``dress" category. Baseline AP scores: 31.5 accuracy, 46.7 reasonableness. RAF model AP scores: 29.2 accuracy, 60.5 reasonableness.}
    \label{fig:supp_example_less_busy}
\end{figure*}

\begin{figure*}
    \centering
    \includegraphics[width=0.99\textwidth]{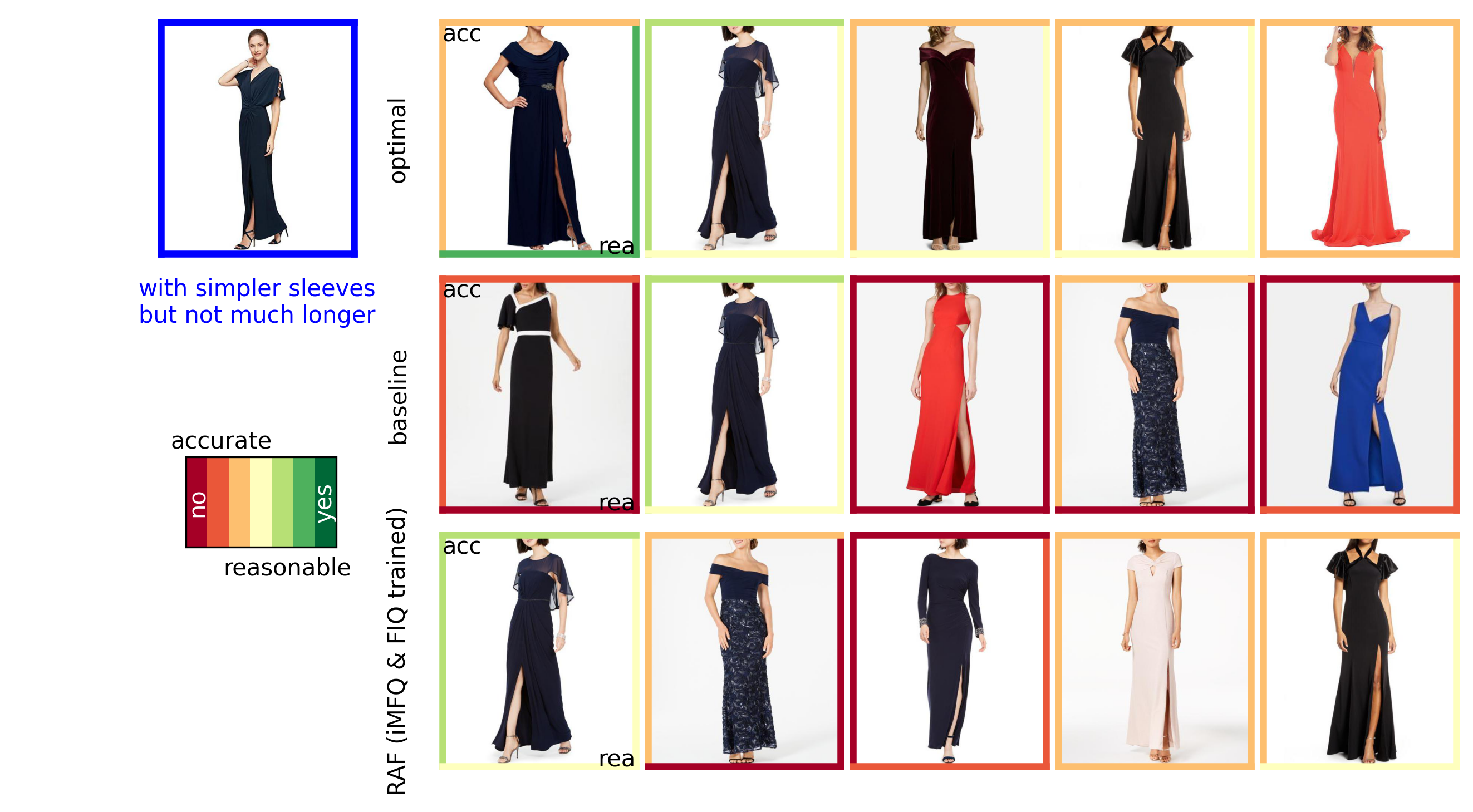}
    \caption{From ``gown" category. Baseline AP scores: 20.0 accuracy, 49.1 reasonableness. RAF model AP scores: 100.0 accuracy, 77.7 reasonableness.}
    \label{fig:supp_example_with_simpler}
\end{figure*}

\begin{figure*}
    \centering
    \includegraphics[width=0.99\textwidth]{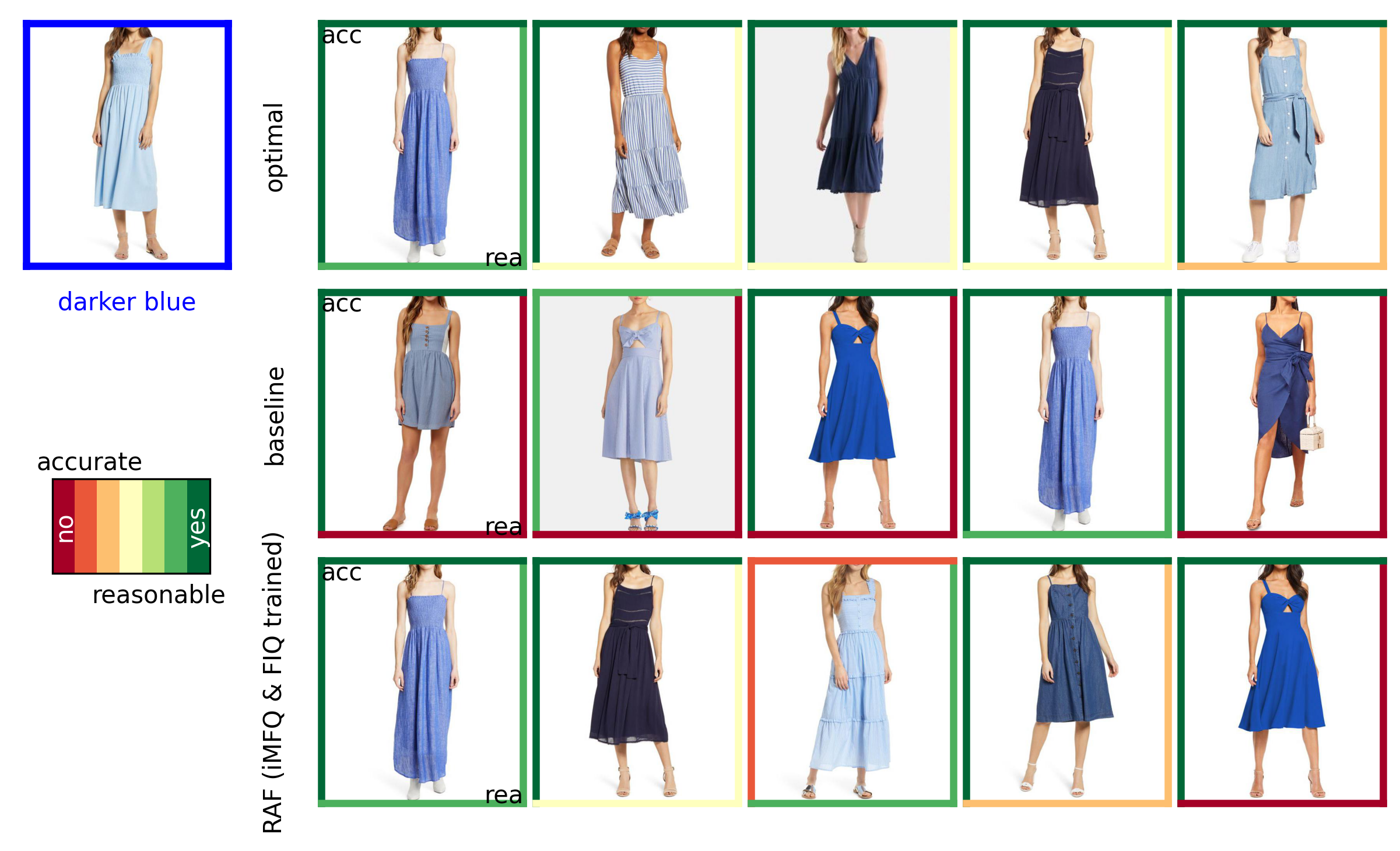}
    \caption{From ``sundress" category. Baseline AP scores: 85.1 accuracy, 42.2 reasonableness. RAF model scores: 86.0 accuracy, 51.3 reasonableness.}
    \label{fig:supp_example_darker_blue}
\end{figure*}

Figures~\ref{fig:supp_example_longer}-\ref{fig:supp_example_darker_blue} show more examples as in Figure~\ref{fig:cfq_example}.

\section{Full results}\label{apx:full-results}
Table \ref{tab:full_results} collects all the experiments presented in section \ref{sec:experiments} with all the metrics used there plus two additional metrics:
\begin{itemize}
    \item We compute mAP for held-out iMFQ data. A response image is considered correct if its attributes match the target attributes, which are the query image's attributes modified according to the caption. 
    \item The normalized discounted cumulative gain (nDCG) for CFQ is a ranking metric that uses the graded judgment scores without thresholding. For relevance scores $r_i$ for the $C$ catalog images and a ranking $\rho$, DCG is
\begin{equation}
    \text{DCG}(\rho) = \sum_{j=1}^{C} \frac{r_{\rho(j)}}{\log_2(\rho(j) + 1)}
\end{equation}
and this is divided by its optimal value to compute nDCG. Here the relevance scores are the sum of the ``accuracy" and ``reasonableness" scores described in section \ref{sec:metrics}, plus a constant 2 to make all $r$ non-negative.
\end{itemize} 
 
\begin{table*}
    \centering
\begin{tabular}{ll|l|lllll|l}
\toprule
training & & FIQ & \multicolumn{4}{c}{CFQ mAP sores} & CFQ & iMFQ \\
data & fusion &      score &          accurate &          reasonable &     relevant &      img reas &             nDCG &         mAP \\
\midrule
N/A & random           &   $0.6^{\pm0.1}$ &  $53.2^{\pm0.7}$ &  $40.1^{\pm0.7}$ &  $22.5^{\pm0.8}$ &  $39.8^{\pm1.0}$ &  $77.6^{\pm0.4}$ &   $0.2^{\pm0.0}$ \\
CLIP & VA        &  $28.4^{\pm0.0}$ &  $59.2^{\pm0.0}$ &  $46.7^{\pm0.0}$ &  $28.4^{\pm0.0}$ &  $53.1^{\pm0.0}$ &  $81.4^{\pm0.0}$ &   $3.7^{\pm0.0}$ \\
CLIP &  image only  &   $8.9^{\pm0.0}$ &  $53.4^{\pm0.0}$ &  $53.1^{\pm0.0}$ &  $29.6^{\pm0.0}$ &  $53.1^{\pm0.0}$ &  $80.5^{\pm0.0}$ &   $3.9^{\pm0.0}$ \\
CLIP &  text only  &  $22.7^{\pm0.0}$ &  $59.8^{\pm0.0}$ &  $40.5^{\pm0.0}$ &  $24.1^{\pm0.0}$ &  - &  $80.5^{\pm0.0}$ &   $1.1^{\pm0.0}$ \\
+ FIQ & image only   &   $9.9^{\pm1.1}$ &  $53.9^{\pm0.5}$ &  $53.8^{\pm0.9}$ &  $30.5^{\pm1.1}$ &  $53.8^{\pm0.9}$ &  $81.0^{\pm0.6}$ &   $4.3^{\pm0.5}$ \\
+ FIQ & text only    &  $38.4^{\pm0.1}$ &  $66.1^{\pm0.4}$ &  $44.0^{\pm0.2}$ &  $30.2^{\pm0.4}$ &  - &  $83.3^{\pm0.2}$ &   $1.6^{\pm0.0}$ \\
+ FIQ & TIRG        &  $44.4^{\pm0.2}$ &  $64.7^{\pm0.5}$ &  $53.3^{\pm1.0}$ &  $36.3^{\pm1.4}$ &  $53.8^{\pm0.4}$ &  $84.9^{\pm0.3}$ &   $6.5^{\pm0.6}$ \\
+ FIQ & AF           &  $46.0^{\pm0.3}$ &  $59.0^{\pm1.1}$ &  $\mathbf{55.7}^{\pm0.5}$ &  $34.7^{\pm1.1}$ &  $54.9^{\pm0.4}$ &  $83.3^{\pm0.5}$ &   $6.8^{\pm0.2}$ \\
+ FIQ  & VA           &  $48.1^{\pm0.3}$ &  $66.8^{\pm0.3}$ &  $52.9^{\pm0.3}$ &  $37.4^{\pm0.3}$ &  $56.5^{\pm0.2}$ &  $85.4^{\pm0.2}$ &   $8.5^{\pm0.1}$ \\
+ FIQ & RAF $\alpha=1$        &  $46.4^{\pm0.5}$ &  $59.6^{\pm0.7}$ &  $55.7^{\pm0.6}$ &  $34.9^{\pm0.5}$ &  $55.0^{\pm0.6}$ &  $83.5^{\pm0.3}$ &   $7.0^{\pm0.2}$ \\
+ FIQ & RAF $\alpha=0.01$          &  $49.7^{\pm0.3}$ &  $\mathbf{67.3}^{\pm0.2}$ &  $53.0^{\pm0.6}$ &  $\mathbf{38.0}^{\pm0.5}$ &  $\mathbf{57.6}^{\pm0.2}$ &  $\mathbf{85.6}^{\pm0.1}$ &   $8.2^{\pm0.1}$ \\
+ iMFQ  & VA        &  $24.9^{\pm0.1}$ &  $60.5^{\pm0.9}$ &  $53.2^{\pm0.3}$ &  $33.2^{\pm0.9}$ &  $\mathbf{55.7}^{\pm0.3}$ &  $84.0^{\pm0.2}$ &  $14.8^{\pm0.1}$ \\
+ iMFQ & RAF        &  $24.4^{\pm0.6}$ &  $60.0^{\pm0.7}$ &  $53.9^{\pm0.4}$ &  $33.4^{\pm0.8}$ &  $56.0^{\pm0.2}$ &  $84.1^{\pm0.4}$ &  $\mathbf{15.1}^{\pm0.1}$ \\
+ iMFQ + FIQ & VA     &  $49.9^{\pm0.3}$ &  $62.9^{\pm0.6}$ &  $\mathbf{55.7}^{\pm0.3}$ &  $37.1^{\pm0.6}$ &  $56.7^{\pm0.3}$ &  $85.0^{\pm0.3}$ &  $12.3^{\pm0.2}$ \\
+ iMFQ + FIQ & RAF     &  $\mathbf{50.2}^{\pm0.3}$ &  $63.9^{\pm0.5}$ &  $55.6^{\pm0.1}$ &  $\mathbf{38.0}^{\pm0.4}$ &  $57.2^{\pm0.0}$ &  $85.4^{\pm0.2}$ &  $12.6^{\pm0.2}$ \\
+ scramble & VA         &   $3.1^{\pm0.7}$ &  $53.1^{\pm1.0}$ &  $45.4^{\pm1.4}$ &  $25.4^{\pm0.8}$ &  $53.1^{\pm0.0}$ &  $79.3^{\pm0.9}$ &   $1.8^{\pm0.2}$ \\
+ mismatch & VA         &   $2.6^{\pm0.0}$ &  $51.3^{\pm0.0}$ &  $40.9^{\pm0.0}$ &  $22.2^{\pm0.0}$ &  $53.1^{\pm0.0}$ &  $76.5^{\pm0.0}$ &   $1.6^{\pm0.0}$ \\
+ scramble + FIQ & VA &  $35.6^{\pm0.3}$ &  $61.5^{\pm3.5}$ &  $48.1^{\pm1.3}$ &  $31.9^{\pm3.2}$ &  $54.6^{\pm0.6}$ &  $83.0^{\pm1.7}$ &   $3.8^{\pm0.6}$ \\
+ mismatch + FIQ & VA &  $32.2^{\pm0.4}$ &  $60.5^{\pm1.4}$ &  $49.3^{\pm1.1}$ &  $32.1^{\pm1.7}$ &  $54.6^{\pm0.4}$ &  $82.9^{\pm0.3}$ &   $4.4^{\pm0.3}$ \\
\bottomrule
\end{tabular}
    \caption{Full results for all experiments on fashion datasets presented in the main text. In the training data column, + indicates sequential training (or scrambling/mismatching the CLIP-trained modules); an initial + means CLIP +. In the fusion column, ``random" indicates a model that outputs random embeddings regardless of input; ``image" ignores the text input; and ``text" ignores the query image (note that the catalog images are still used). The ``img reas" column is the same metric as the ``reasonable" column but the model is not given any text input.}
    \label{tab:full_results}
\end{table*}

\subsection{Fashion IQ score breakdown}

Prior work usually reports R@10 and R@50 for each of the three categories in Fashion IQ as was done in the paper introducing the dataset\cite{guo2019fashion}. The different scores are highly correlated with each other across models, so for space and simplicity we reported only the single summary score in the main text. In Table \ref{tab:fiq_breakdown} we report the finer metrics for direct comparison with other work.

\begin{table*}[]
    \centering
\begin{tabular}{l|ll|ll|ll|ll}
 & \multicolumn{2}{c|}{Dress} & \multicolumn{2}{c|}{Top\&Tee} & \multicolumn{2}{c|}{Shirt} & \multicolumn{2}{c}{Average}\\
{Method} &        R@10 &        R@50 &       R@10 &       R@50 &        R@10 &        R@50 & R@10 & R@50 \\
\hline
JVSM\cite{chen2020learning} & 10.7 & 25.9 & 13.0 & 26.9 & 12.0 & 27.1 & 11.9 & 26.6  \\
% below from CoSMo paper except CIRPLANT
Relationship\cite{santoro2017simple} & 15.44 & 38.08 & 21.10 & 44.77 & 18.33 & 38.63 & 18.29 & 40.49  \\
MRN\cite{kim2016multimodal} & 12.32 & 32.18 & 18.11 & 36.33 & 15.88 & 34.33 & 15.44 & 34.28  \\
FiLM\cite{perez2018film} & 14.23 & 33.34 & 17.30 & 37.68 & 15.04 & 34.09 & 15.52 & 35.04  \\
TIRG\cite{vo2019composing} & 14.87 & 34.66 & 18.26 & 37.89 & 19.08 & 39.62 & 17.40 & 37.39  \\  % from CoSMo paper, not optimal
CIRPLANT\cite{liu2021image} & 17.45 & 40.41 & 21.64 & 45.38 & 17.53 & 38.81 & 18.87 & 41.53  \\
VAL\cite{Chen_2020_CVPR} & 21.12 & 42.19 & 25.64 & 49.49 & 21.03 & 43.44 & 22.60 & 45.04  \\
CoSMo\cite{Lee_2021_CVPR} & 25.64$^{\pm 0.21}$ & 50.30$^{\pm 0.10}$ & 29.21$^{\pm 0.12}$ & 57.46$^{\pm 0.16}$ & 24.90$^{\pm 0.25}$ & 49.18$^{\pm 0.27}$ & 26.58 & 52.31 \\
% not from CoSMO anymore
RTIC-GCN\cite{shin2021rtic} & 29.15 & 54.04 & 31.61 & 57.98 & 23.79 & 47.25 & 21.18 & 53.09 \\
DCNet\cite{kim2021dual} & 28.95 & 56.07 & 30.44 & 58.29 & 23.95 & 47.30 & 27.78 & 53.89 \\
% didn't see before submitting to cvpr
CLVC-Net & 29.85 & 56.47 & 33.50 & 64.00 & 28.75 & 54.76 & 30.70 & 58.41 \\
\hline
CLIP \cite{clip} VA     &  $16.5$ &  $35.2$ &  $21.7$ &  $41.9$ &  $19.5$ &  $35.7$ & 19.2 & 37.6 \\
CLIP VA + FIQ        &  $31.1^{\pm0.4}$ &  $57.1^{\pm0.5}$ &  $39.5^{\pm0.8}$ &  $67.2^{\pm0.3}$ &  $34.4^{\pm0.3}$ &  $59.7^{\pm0.3}$ & 35.0 & 61.3 \\
+ RAF (Ours) &  $\mathbf{32.9}^{\pm0.6}$ &  $\mathbf{59.9}^{\pm0.2}$ &  $\mathbf{42.2}^{\pm0.4}$ &  $\mathbf{67.8}^{\pm0.2}$ &  $\mathbf{37.0}^{\pm0.9}$ &  $\mathbf{61.6}^{\pm0.4}$ & \textbf{37.4} & \textbf{63.1} \\
\hline
\end{tabular}
    \caption{Fashion IQ metrics on the same models as in Table \ref{tab:comparison_fiq} in the main text.}
    \label{tab:fiq_breakdown}
\end{table*}

\section{Training details}
Models were trained with batch-wise softmax cross-entropy\cite{sohn2016improved} with a learned inverse-temperature parameter, using Adam~\cite{kingma2014adam} for 14 FIQ epochs and/or 3 iMFQ epochs on one NVIDIA V100 GPU. This procedure clearly left the ``scrambled" or ``mismatched" models undertrained, so we trained these models for 42 epochs. The learning rate was set to $10^{-6}$ for VA models and dropped by a factor of 10 after half the training epochs for FIQ and after each epoch for iMFQ. For models with fusion modules, the associated weights were randomly initialized and given 10x the learning rate of the pretrained parameters. 

For FIQ only, we used simple image augmentations: horizontal flips, resize and crop with scales 0.8-1.0 and ratios 0.75-1.3, and Gaussian pixel noise with standard deviation 0.1.

\section{CFQ metrics}

\begin{figure*}
    \centering
    \includegraphics[width=0.95\textwidth]{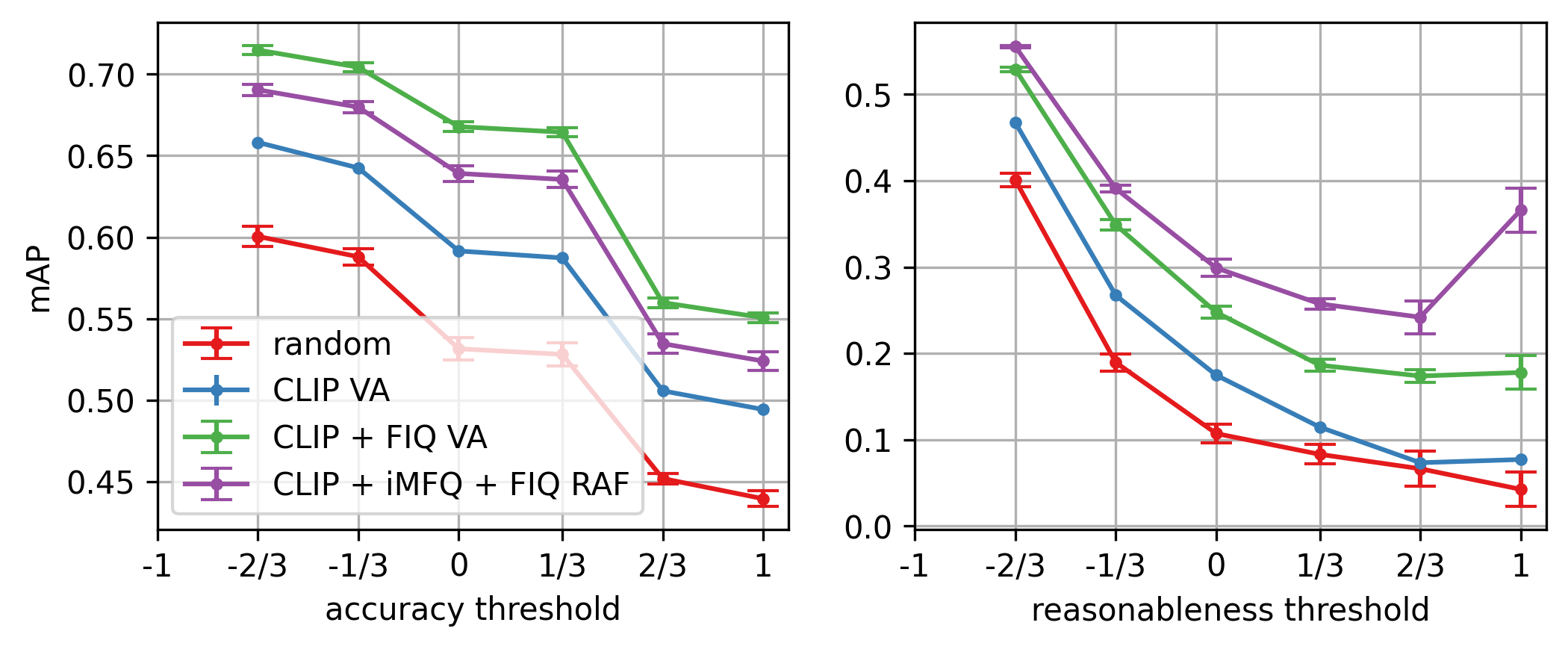}
    \caption{How reported mAP metrics would change if we had used different thresholds for the judgments. Several representative model results are shown. For the ``random" baseline, we plotted mean and standard deviation (error bars) over five instances.}
    \label{fig:thresholds}
\end{figure*}

Our mAP metrics for CFQ depend on a choice of threshold for what counts as a positive judgment for each question considering the three annotators' responses. Figure~\ref{fig:thresholds} shows results for several models on these metrics as we vary the threshold. When not otherwise specified, our accuracy mAP results use threshold 0 and our reasonableness mAP results use threshold -2/3 (meaning ``not reasonable" must be a unanimous judgment or we count it as reasonable). Recall that the overall relevance judgment is the logical AND of the accuracy and reasonableness judgments.

We observed that while the base rate of the mAP metrics depends strongly on the thresholds (since chance performance is the fraction of positives), varying the threshold does not usually change the order in which models score. The unthresholded judgments still inform the nDCG metric, which we report in appendix \ref{apx:full-results}.

\end{document}